\documentclass[conference]{IEEEtran}
\IEEEoverridecommandlockouts
\usepackage{cite}
\usepackage{amsmath,amssymb,amsfonts}
\usepackage{algorithmic}
\usepackage{graphicx}
\usepackage{textcomp}
\usepackage{xcolor}
\usepackage{hyperref}
\def\BibTeX{{\rm B\kern-.05em{\sc i\kern-.025em b}\kern-.08em
    T\kern-.1667em\lower.7ex\hbox{E}\kern-.125emX}}

\definecolor{darkpastelgreen}{rgb}{0.01, 0.75, 0.24}

\begin{document}

\title{AutoML for neuromorphic computing and application-driven co-design: asynchronous, massively parallel optimization of spiking architectures 
\thanks{This work was supported by DOE ASCR and BES Microelectronics Threadwork. This material is based upon work supported by the U.S. Department of Energy, Office of Science, under contract number DE-AC02-06CH11357.}
}

\author{\IEEEauthorblockN{Angel Yanguas-Gil}
\IEEEauthorblockA{\textit{Applied Materials Division} \\
\textit{Argonne National Laboratory}\\
Lemont, IL, USA \\
ayg@anl.gov}
\and
\IEEEauthorblockN{Sandeep Madireddy}
\IEEEauthorblockA{\textit{Mathematics and Computer Science Division} \\
\textit{Argonne National Laboratory}\\
Lemont, IL, USA \\
smadireddy@anl.gov
}
}


\maketitle

\begin{abstract}
In this work we have extended AutoML inspired approaches to the exploration and optimization of neuromorphic architectures. Through the integration of a parallel asynchronous model-based search approach with a simulation framework to simulate
spiking architectures, we are able to efficiently explore
the configuration space of neuromorphic architectures and identify the subset of conditions leading to the highest performance in a targeted application. We have demonstrated this approach on an exemplar case of real time, on-chip learning application. Our results indicate that we can effectively use optimization approaches to optimize complex architectures, therefore providing a viable pathway towards application-driven codesign.
\end{abstract}


\section{Introduction}

The vast design space in neuromorphic computing is one of the challenges that can slow down the deployment of optimized 
architectures. This space encompasses multiple aspects including
network architecture, neuron models, synaptic plasticity mechanisms,
and input encoding. This problem is compounded if we also consider potential hardware implementations, which span from digital ASIC to hybrid digital/analog design to a multiplicity of emergent devices.

Some of that complexity is common to traditional hardware design. Other aspects however are central to the nature of neuromorphic computing, and we can find counterparts in biological systems: beyond
the simplified neuron models, there are hundreds of neuropeptides, neuromodulators, receptors, and signaling mechanisms that have
been selected for their ability to enable complex functionality 
of the central nervous system. Disruptions in this space often lead to
dramatic drops in performance, as showcased in the literature (See for instance Ref.
\cite{KOOB20041515}).

In recent years, there has been an increasing interest in the use of machine learning approaches for hardware design, using techniques such as reinforcement learning for automatic placement during the chip design stage.\cite{Zhang_2022} One possible approach is to use automatic machine learning (AutoML) approaches used in deep learning as an inspiration to tackle this issue in the context of neuromorphic computing. Steps such as input encoding, neuron and synapse model, synaptic plasticity rule, and network architecture selection, and hyperparameter optimization can potentially be automated. In the context of neuromorphic architectures, this process would help us quickly identify the subset of the design space that is more promising for a specific application or that would capture the benefits and behavior
of emergent devices. This is a crucial step to 
bring neuromorphic computing to the mainstream, 
particularly to resource-constrained, SWaP-C scenarios.

The same approach can be used to better understand how to adapt spiking neural networks (SNNs) to specific workflows.
A strategy often used in the literature is to hide that complexity from the user of neuromorphic hardware, for instance by operating at higher levels of 
abstraction.\cite{Nengo, Whetstone} While useful in their contexts,  this strategy doesn't work when we are trying to leverage
the unique properties of spiking neural networks. In that case, access to the lower level details
of the SNN is crucial. For instance, the two generations of Intel's Loihi,\cite{Loihi} are designed to be flexible and enable a wide range of neurons and synaptic plasticity models. The ability to effectively search for optimal configurations within this vast design space can help accelerate the development of novel algorithms in existing
neuromorphic hardware.

In this work, we explore such an approach: we have demonstrated the ability to carry out massively parallel configuration searches to identify optimal regions in the design space of neuromorphic architectures, using their performance on specific tasks as the optimization target. To this end, we have developed spikelearn, a lightweight framework that can simulate spiking neural networks in a wide range of non-trivial tasks, including on-chip learning with a wide range synaptic plasticity rules spanning from neuro-inspired to memristor-based. This framework allows us to run non-trivial tasks with millions of synapses using a single core on leadership computing machines, opening up the ability to do configuration searches at the exascale. It can also operate on smaller machines relying on just a few cores. To efficiently search the design space of neuromorphic architectures, we have coupled this framework with a parallel asynchronous model-based search approach. As an exemplar, we have explored two configuration spaces focused on online learning using a spiking analog of a covariance learning rule. The optimization run are carried out under stringent conditions both in terms of data availability and number of spikes per sample, which allow us to explore the transfer of these configurations to other conditions.

\section{Methodology}

\subsection{Simulation framework: streamnet and spikelearn}

One of the challenges of applying AutoML to neuromorphic architectures is that any simulation framework used needs to satisfy the following requirements: 1) it has to be able to carry out workflows at the scale of the relevant applications. This may involve running hundreds of thousands of samples in some cases. 2) They have to allow the simulation of a wide range of neurons, architectures, and synaptic plasticity rules. 3) They should be able to incorporate behaviors representative of novel devices which go beyond those typically found in spiking simulators. 4) They should be able to run effectively in high performance computing machines.

\begin{figure}[thbp]
\centerline{\includegraphics[width=8cm]{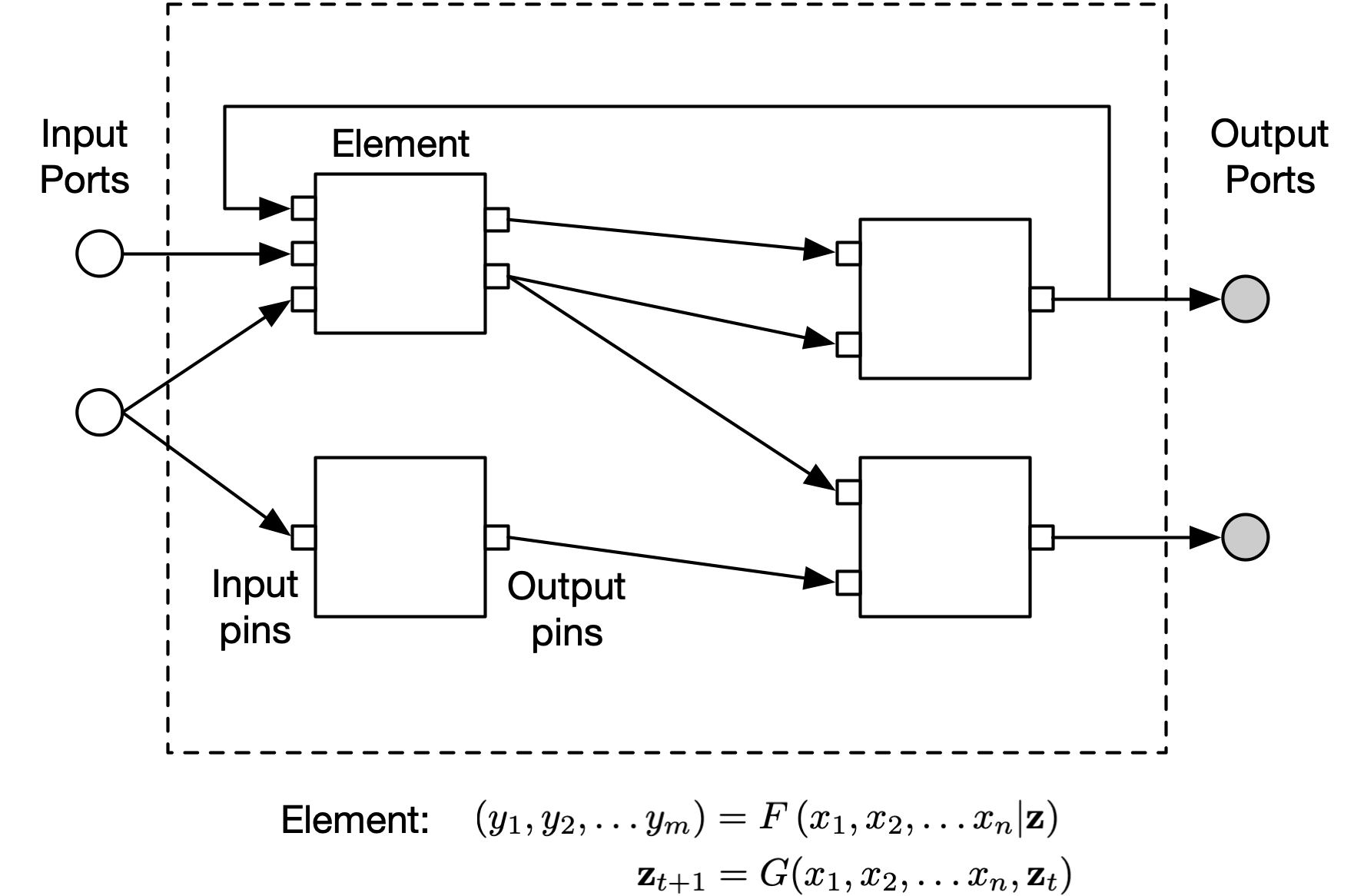}}
\caption{Streamnet is the data structure at the core of the simulation framework: a streamnet represents a directed graph with labeled nodes where inputs to each node are restricted to an indegree of one. This structure enables the simulation of heterogeneous networks where all components advance at a common time step. Computation within each node is hidden from the rest of the network, which allows us to integrate SPICE, HDL, Tensorflow/Pytorch, SNN, and physics-based components in a modular fashion.}
\label{fig:streamnet}
\end{figure}

In order to incorporate these four requirements we have developed a lightweight tool based on streamnets (Fig. \ref{fig:streamnet}). A \emph{streamnet} is a data structure comprising a graph of \emph{Elements} with two ordered sets of \emph{input nodes} and \emph{output nodes}. A key restriction with respect to what would amount to a netlist is that each input node can be connected to just one output node (indegree of one). This restriction allows us to implement a simple execution model where at each time step, each element pulls the output from its corresponding output node and uses it to compute the next step. In order to interact with the outer world, a streamnet also defines a list of input ports and output ports.

The streamnet structure is ideally suited to the simulation of neural networks, since: 1) the exact computation carried out within each element is hidden from the streamnet. Consequently, we can integrate multiple simulation tools as long as they can be encapsulated within a single Element; 2) the time step $\Delta t$ can be chosen to represent the minimum propagation delay in the network (axon delay); and 3) It allows swapping elements without breaking the simulation as long as they share the same interface. 4) A streamnet is also an element. This adds a layer of composability that is very similar to that found in machine learning frameworks such as PyTorch.

Spikelearn implements spiking neural networks using the streamnet data structure as the backbone. Spikelearn implements a series of synaptic plasticity rules that are meant to be a superset of that implemented in Loihi,\cite{Loihi} extending it to incorporate ternary synaptic plasticity rules that use a modulatory signal to regulate learning potentially in a synapse-by-synapse fashion, as well as synaptic plasticity models based on the VTEAM memristor model.\cite{VTEAM}
Spikelearn has been released and can be found at \url{https://github.com/spikelearn/spikelearn}

\subsection{Massively parallel optimization with deepHyper}

In this work, we seek to optimize neuromorphic architectures across the choice of neuron models, 
synaptic models, and synaptic plasticity models and the tunable parameters 
in each of these models. Formally, the parameter space can be written 
as $x = (x_{\mathcal{I}},x_{{\mathcal{R}}},x_{{\mathcal{C}}}$), where 
$\mathcal{I},\mathcal{R},\mathcal{C}$ respectively denote integer, 
continuous, and categorical parameters. The number of neurons in a given layer, the leakage time of a spiking neuron, and the type of synaptic plasticity rule are examples of each of these in the context of this work. The resulting optimization problem, 
which seeks to maximize the accuracy of the model as a function of 
this mixed space, is, in fact, a non-convex black-box mixed-integer 
nonlinear optimization~\cite{burer2012non}.  

Due to the heterogeneity of the search space and the black-box nature of the simulation framework and objective function, we adopted a parallel asynchronous 
model-based search approach (AMBS)~\cite{Balaprakash_DH_2018} 
to find the high-performing parameter configurations. 
The AMBS approach consists of sampling-based Bayesian
optimization that samples a number of parameter configurations 
from this space $x$ and progressively fits a surrogate model 
(\emph{random forests} in our case) over the parameter configurations' 
accuracy metric space. This surrogate model is updated asynchronously 
as new configurations are evaluated by parallel processes, which are 
then used to obtain configurations that will be evaluated in the next 
iteration. 

We adopt \emph{random forests} over other generic choices
such as \emph{Gaussian process} due to the ability of the former to build
surrogate models effectively in a mixed space. It does so by building
multiple decision trees and using bootstrap aggregation (or bagging) 
to combine them to produce a model with better predictive accuracy 
and lower variance. In addition, the acquisition function plays an 
important role in maintaining the exploration-exploitation balance 
during the search. With AMBS we adopt the {\it lower confidence bound} 
(LCB) acquisition function.

Since AMBS enables asynchronous search, a large number of hyperparameter 
configurations can be evaluated in parallel and inform the surrogate 
modeling, which is crucial for search over high-dimensional and complex 
mixed parameter spaces. The complexities of running such large numbers of evaluations 
on high-performance computing systems are handled by tight integration with
workflow management frameworks such as Ray~\cite{moritz2018ray} and MPI~\cite{gropp1999using}.

\section{Experiments}

In order to explore the ability to efficiently search and optimize neuromorphic architectures we have used
online, on-chip learning as an exemplar. 

\subsection{Task structure and optimization metrics}

The tasks explored in this work are all instances
of stream learning, where the architecture is
learning directly from a stream of inputs:
starting from random synaptic weights, learning takes
place in real time and, after
a predetermined number of samples, we evaluate the 
achieved accuracy in a separate stream learning assay where learning is disabled. These tasks can be viewed as
\emph{metalearning experiments}, where we are evaluating
and optimizing the system's ability to learn.

In the supervised case, the neuromorphic architecture receives class information as additional input and
the final classification
accuracy is used as target for optimization.
In the unsupervised case, we follow the same task structure, except that we use an l$_2$ metric
to compute the distance between the input and the projection in the feature space defined by the synaptic weights.


\subsection{Neuron model and synaptic plasticity rule}

We have considered leaky integrate and fire neurons given by the following equations:
\begin{eqnarray}
v(n+1) & = & \left(1-s(n)\right)\left(v(n) e^{-1/\tau} + x(n)\right) \\
s(n+1) & = & \mathrm{H}\left(v(n)-v_\mathbf{th}\right)
\end{eqnarray}
Here $v(n)$ is the membrane potential, $s(n)$ is the output spike of the neuron, $x(n)$ represent the sum of all synaptic inputs at $n$, 
$\mathrm{H}\left(\cdot\right)$ is the Heaviside function, $\tau$ is
the characteristic decay time of the membrane potential, and $v_\mathrm{th}$ is the firing threshold voltage of the neuron. This
model is analogous to Loihi's model of a leaky integrate and fire
neuron.\cite{Loihi} All synapses can be either pass through or low pass.

In supervised experiments we have used
a ternary, modulated non-Hebbian synaptic plasticity
rule that is inspired on experimental neuroscience.\cite{Aso_MB_2014} 
This rule considers presynaptic, postsynaptic, and modulatory traces ($t_e$, $t_o$, $t_m$), where each trace is computed as: $t(n) = a t(n-1) + b s(n)$, so that the change in synaptic weight $\Delta W$ is given by:
\begin{equation}
    \label{MSErule}
    \Delta W = l_r t_e (R_0 t_m - t_o)
\end{equation}
Here $l_r$ represents the learning rate, and $R_0$ is a proportionality constant relating the modulatory input and the post-synaptic output of the neuron. 
Finally, we can further impose upper and lower bounds to the synaptic weights. These can be defined between $[0, W_\mathrm{max}]$ or$[-W_\mathrm{max}, W_\mathrm{max}]$ depending on the nature of the synapse (excitatory/inhibitory vs hybrid).

This model can be viewed in two ways: it resembles some of the covariance rate models,\cite{dayan2005theoretical} where the modulatory signal is a moving threshold and it tracks the anticovariance instead. It can also be viewed as an spiking counterpart of a mean square error loss function. Consequently, in this work we will refer to this rule as the MSE rule.

\subsection{Test cases}

\subsubsection{Learning rule optimization in a shallow network}

The shallow model consists simply of a layer of neurons each receiving direct inputs from the dataset. Inputs are encoded as Poisson spike trains. This shallow model will help us focus on the synaptic plasticity rule and evaluate the fraction of the design space that 
is conducive to online learning.
The design space for optimization is described in Table \ref{tab:shallow}.

\begin{table}[htbp]
\caption{Design space for the shallow case}
\begin{center}
\begin{tabular}{|c|c|c|c|}
\hline
Component & Parameter & Explanation & Range \\
\hline
output neuron & $\tau$ & Leakage time & 1-8 \\
\hline
 & $\tau$ & Low-pass filter of input spk. & 0.1-4 \\
\cline{2-4}
 & $\tau_m$ & Low-pass filter of mod. spk. & 0.1-4 \\
\cline{2-4}
 & $t_e$: $(a,b)$ & Presynaptic trace & 0.05-1 \\
\cline{2-4}
MSE syn. & $t_o$: $(a,b)$ & Postsynaptic trace & 0.05-1 \\
\cline{2-4}
 & $t_m$: $(a,b)$ & Modulatory trace & 0.05-1 \\
\cline{2-4}
 & $R_0$ & Weight of modulatory trace & 0.1-4 \\
\cline{2-4}
 & $l_r$ & Learning rate & $10^{-4}$-0.5 \\
\cline{2-4}
 & $W_\mathrm{lim}$ & Weight limit & 0.01-0.5 \\
\hline
\end{tabular}
\label{tab:shallow}
\end{center}
\end{table}

\begin{figure}[thbp]
\centerline{\includegraphics[width=8cm]{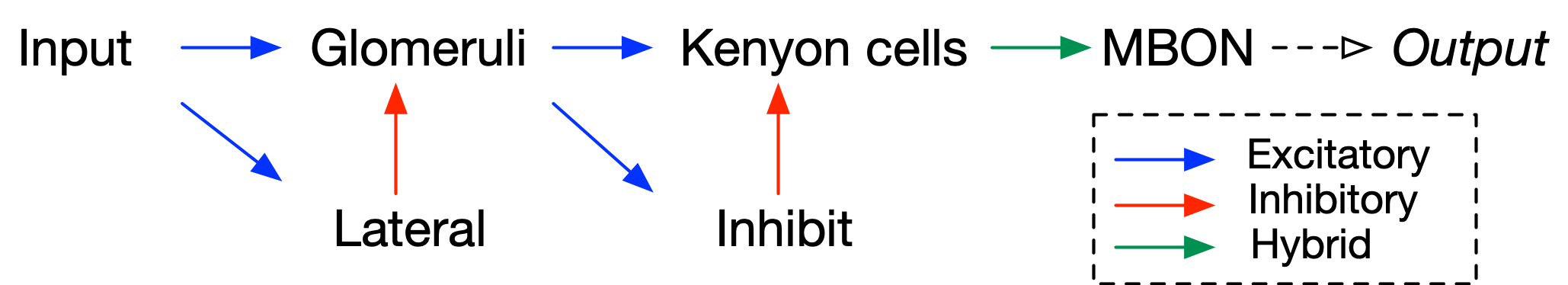}}
\caption{Architecture for the complex case, inspired on the
olfactive system of the insect brain}
\label{fig:mb}
\end{figure}

\subsubsection{Optimization of a complex network}

The complex model implements a deeper architecture that is inspired in the olfactive system of the insect brain.\cite{Aso_MB_2014,Strausfeld_2012} As shown in Figure \ref{fig:mb}, Inputs are passed to the glomeruli layer and to a lateral neurons which inhibits the neurons in the glomeruli. These glomeruli then project into a sparse layer (Kenyon cells) before densely fan in into a set of output neurons. The glomeruli also project into an inhibitory neuron that inhibits the whole population of Kenyon cells. Finally, a modulatory signal is encoded by a set of modulatory neurons that mimic the dopamine clusters that innervate the synapses between Kenyon cells and the output neurons. These modulatory neurons provide the supervised information for each of the categories, as in the shallow case. Learning takes place at synapses between the mushroom body and the output neurons, for a total of 50,000 active synapses in the case of MNIST.

The resulting design space is shown in Table \ref{tab:complex}, where we include both the parameters and the range considered during the configuration search.

\begin{table}[htbp]
\caption{Design space for the complex case}
\begin{center}
\begin{tabular}{|c|c|c|c|}
\hline
Component & Parameter & Explanation & Range \\
\hline
glom. neuron & $\tau$ & Leakage time & 1-8 \\
\hline
lat. neuron & $\tau$ & Leakage time & 1-8 \\
\hline
kcell. neuron & $\tau$ & Leakage time & 1-8 \\
\hline
inhib. neuron & $\tau$ & Leakage time & 1-8 \\
\hline
mbon neuron & $\tau$ & Leakage time & 1-8 \\
\hline
inp2glom syn. & $W_0$ & Synaptic weight & 2-10 \\
\hline
inp2lat syn. & $W_0$ & Synaptic weight & 0-0.5 \\
\hline
lat2glom syn. & $W_0$ & Synaptic weight & 0-2 \\
\hline
glom2inh syn. & $W_0$ & Synaptic weight & 0-0.2 \\
\hline
glom2kc syn. & $W_0$ & Synaptic weight & 0-0.2 \\
\cline{2-4}
 & $p_0$ & Connection probability & 0.01-0.05 \\
\hline
 & $\tau$ & Low-pass filter of input spk. & 0.1-4 \\
\hline
 & $\tau_m$ & Low-pass filter of mod. spk. & 0.1-4 \\
\cline{2-4}
 & $t_e$: $(a,b)$ & Presynaptic trace & 0.05-1 \\
\cline{2-4}
kc2mbon syn. & $t_o$: $(a,b)$ & Postsynaptic trace & 0.05-1 \\
\cline{2-4}
 & $t_m$: $(a,b)$ & Modulatory trace & 0.05-1 \\
\cline{2-4}
 & $R_0$ & Weight of modulatory trace & 0.1-4 \\
\cline{2-4}
 & $l_r$ & Learning rate & $10^{-4}$-0.5 \\
\cline{2-4}
 & $W_\mathrm{lim}$ & Weight limit & 0.01-0.5 \\
\hline
\end{tabular}
\label{tab:complex}
\end{center}
\end{table}

\section{Results}

In this work, we optimized the two architectures against the online learning of the MNIST dataset. In each case we streamed
10,000 images through the architectures, which is 1/6 of an epoch, starting from a random configuration. The goal is therefore to identify configurations leading to fast learning, a desirable attribute for online learning scenarios. We use 8 timesteps per sample: this reduces the number of output spikes to 4 in the case of the shallow network, and 6 in the complex case.

\begin{figure}[thbp]
\centerline{\includegraphics[width=8cm]{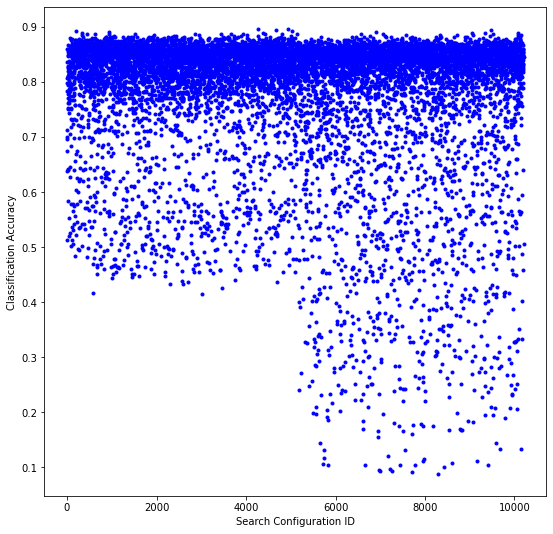}}
\caption{Accuracy on the online learning task obtained during the search of the configuration space for the shallow case. The concentration of data points around the top accuracy values are consistent with the MSE synaptic plasticity rule enabling supervised learning under a wide range of parameters}
\label{fig:shallow}
\end{figure}

In Figure \ref{fig:shallow}  we show the
evolution on task accuracy during a single configuration search using
spikelearn and deepHyper for the shallow case. The search algorithm explored a total of 10,204 configurations. The results obtained indicate that the algorithm could effectively identify a broad region of high accuracy. In this exemplar case, the implication is that the MSE rule described in Section III can be an effective rule for supervised learning in spiking neural networks. More broadly, this validates our methodology for
architecture optimization.

\begin{figure}[thbp]
\centerline{\includegraphics[width=8cm]{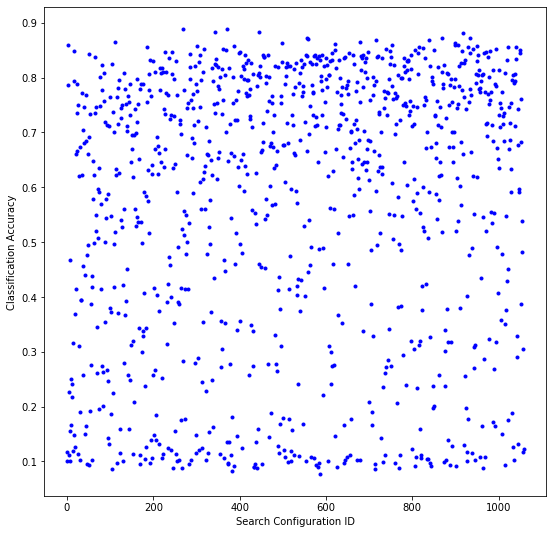}}
\caption{Accuracy on the online learning task obtained during 
a single search of the configuration space for the complex case. The higher dimensional configuration space leads to a more homogeneous distribution of accuracy values, consistent with a smaller subset of the parameter space capable of achieve high accuracy in the online learning task}
\label{fig:complex}
\end{figure}

\begin{figure}[thbp]
\centerline{\includegraphics[width=8cm]{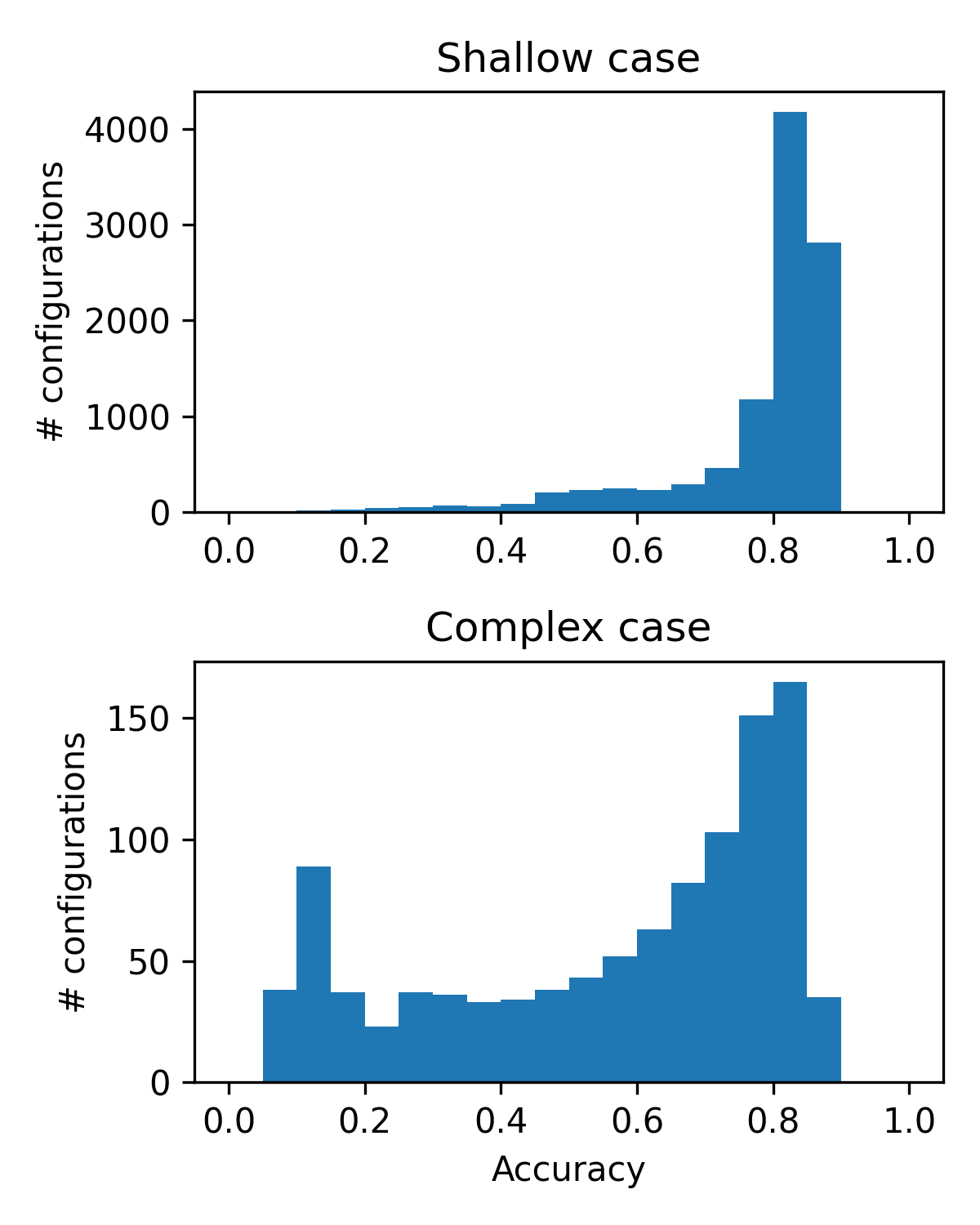}}
\caption{Distribution of task accuracies obtained during a single
configuration search using deep hyper for the shallow and complex cases.}
\label{fig:hist}
\end{figure}

Similarly, we show a snapshot of a single search for the complex case in Figure \ref{fig:complex}, comprising 1,059 configurations. In this case, the data points are more evenly spread across the accuracy range. This is consistent with a narrower region of the search space leading to high accuracy in the online learning task, probably due to the higher dimensionality of the configuration space and the fact that in this complex case we are trying to optimize the parameters for all neurons and synapses concurrently. In Figure \ref{fig:hist} we show a comparison between the distributions of task accuracy achieved during the searches in
the shallow and complex case. The distribution for the shallow case is heavily skewed towards the high accuracy values compared to the complex case.

\begin{figure}[thbp]
\centerline{\includegraphics[width=8cm]{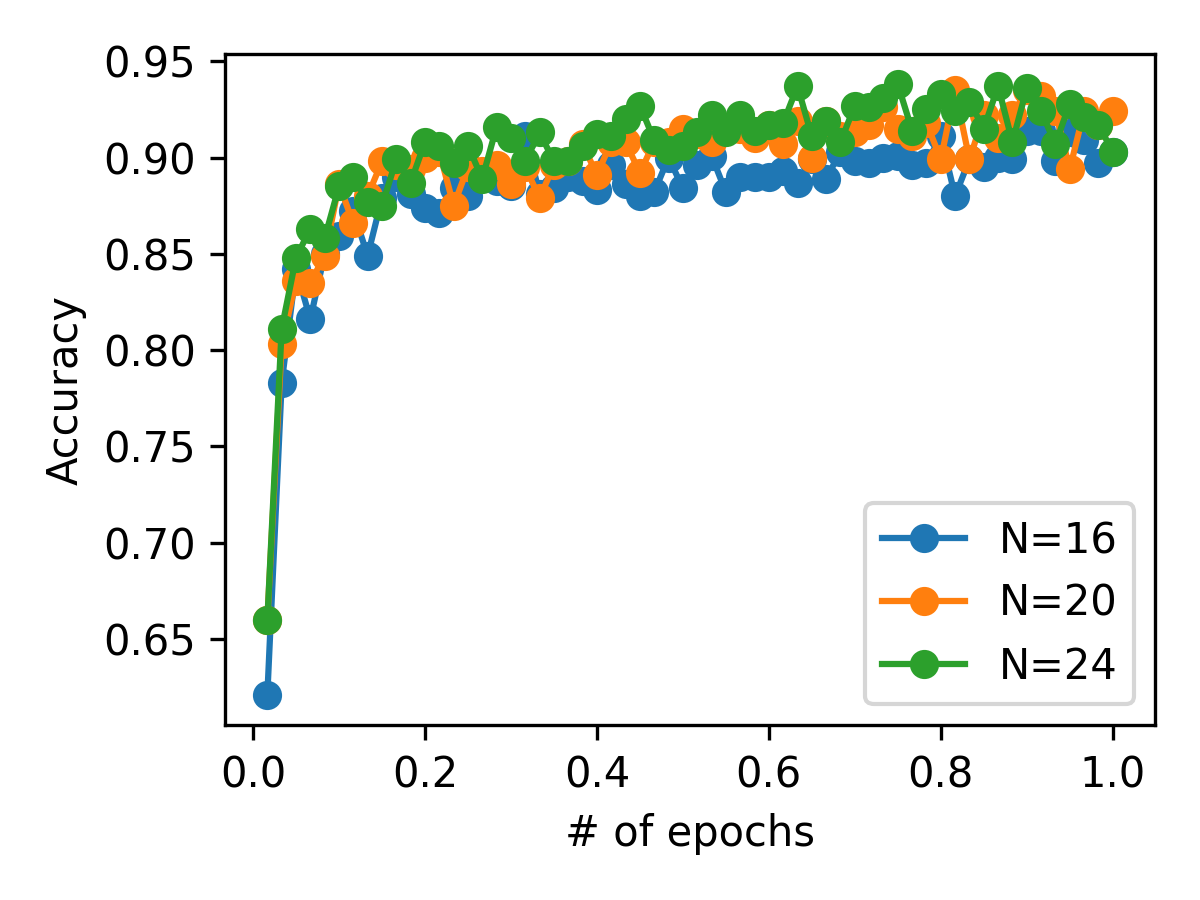}}
\caption{Evolution of classification accuracy during the online learning tasks for different number of steps. When a number of 20 steps is used, the classification accuracy exceed 93\%, which is consistent with the values obtained using stochastic gradient
descent methods.}
\label{fig:transfer}
\end{figure}

Finally, the optimization experiments were carried out using only 10,000 samples and a small number of simulation steps. This may result on accuracy values that are lower than those expected when using longer encodings. In order to understand the robustness of the configurations identified by the search algorithm we carried out experiments where we explored longer tasks both in terms of the number of samples and in the number of spikes. In Figure \ref{fig:transfer}
we show the evolution of task accuracy during the online learning of
the MNIST task for three different numbers of steps per sample: 16, 20, and 24. The results show that for 20 steps and higher, the classification accuracy reaches values close to 93\%. This are consistent with values reported in prior works training spiking neural networks with stochastic gradient descent methods,\cite{ayg_spikes} indicating that the optimization process indeed identifies configurations that are close to the optimal values. This is a necessary step to explore more complex metalearning experiments involving distributions of tasks.

\section{Conclusions}

In this work we have demonstrated the application of AutoML-inspired approaches to the exploration and optimization of neuromorphic architectures. Through the combination of a flexible simulation framework (spikelearn) and a parallel asynchronous model-based search strategy (deepHyper), we have been able to explore the configuration space of shallow and deep spiking neural networks and identify optimal configurations that maximize their performance on a predetermined task. The simulation framework has been designed to be lightweight and seamlessly integrate different types of simulation approaches for each of the components, from HDL to physics-based models and conventional artificial neural networks. As part of this work, we have demonstrated its ability to run in high performance computing environment and leverage the extreme parallelism afforded by these facilities to explore a large number of configurations, more than 10,000 in the shallow case.

Through the exemplar problem, we have been able to identify a novel modulated synaptic plasticity rule for online supervised learning in spiking architectures. Beyond this specific case, the results presented in this work demonstrate that AMBS approaches provide
a viable pathway towards application-driven optimization of neuromorphic architectures. We are currently working to extend the design space to incorporate emergent devices such as memristors and memtransistors into the spiking architectures for high energy and nuclear physics detectors.

Finally, it is important to point out that a key requirement for the development of neuromorphic architectures capable of on-chip learning is the identification of synaptic plasticity rules that are robust and performant against a broad distribution of tasks. The exemplar case explored in this work, while in itself trivial, provides a starting point for more robust metalearning experiments involving neuromorphic architectures in areas such as signal processing or autonomous agents.

\bibliography{references}
\bibliographystyle{plain}

\end{document}